\documentclass[11pt]{article}

\usepackage{graphicx}
\usepackage{booktabs}
\usepackage{multirow}
\usepackage{amsmath}
\usepackage{amssymb}
\usepackage[numbers]{natbib}
\usepackage{hyperref}
\usepackage{geometry}
\usepackage{placeins}
\usepackage{caption}
\usepackage{subcaption}
\usepackage{array}
\usepackage{longtable}
\usepackage{siunitx}
\usepackage{enumitem}
\usepackage{microtype}

\hypersetup{
  colorlinks=true,
  linkcolor=blue,
  citecolor=blue,
  urlcolor=blue
}

\title{A Comparative Study of Custom CNNs, Pre-trained Models, and Transfer Learning Across Multiple Visual Datasets}
\author{Annoor Sharara Akhand \\ University of Dhaka \\ \texttt{sharara99@gmail.com}}
\date{}

\begin{document}
\maketitle

\begin{abstract}
Convolutional Neural Networks (CNNs) are a standard approach for visual recognition due to their capacity to learn hierarchical representations from raw pixels. In practice, practitioners often choose among (i) training a compact custom CNN from scratch, (ii) using a large pre-trained CNN as a fixed feature extractor, and (iii) performing transfer learning via partial or full fine-tuning of a pre-trained backbone. This report presents a controlled comparison of these three paradigms across five real-world image classification datasets spanning road-surface defect recognition, agricultural variety identification, fruit/leaf disease recognition, pedestrian walkway encroachment recognition, and unauthorized vehicle recognition. Models are evaluated using accuracy and macro F1-score, complemented by efficiency metrics including training time per epoch and parameter counts. The results show that transfer learning consistently yields the strongest predictive performance, while the custom CNN provides an attractive efficiency--accuracy trade-off, especially when compute and memory budgets are constrained.
\end{abstract}

\section{Introduction}

Computer vision has undergone a profound transformation over the past decade, largely driven by advances in deep learning and the widespread adoption of Convolutional Neural Networks (CNNs). Unlike traditional vision pipelines that rely on hand-crafted features such as SIFT or HOG, CNNs learn hierarchical feature representations directly from raw pixel data through end-to-end optimization \citep{lecun1998gradient,krizhevsky2012imagenet}. This ability to jointly learn low-level, mid-level, and high-level visual abstractions has enabled CNNs to achieve state-of-the-art performance across a broad spectrum of tasks, including image classification, object detection, semantic segmentation, and visual scene understanding.

The success of CNNs has been further amplified by the availability of large-scale annotated datasets and increased computational power, particularly through GPU acceleration. Benchmark datasets such as ImageNet have played a pivotal role in demonstrating the scalability of deep CNNs and their capacity to generalize across diverse visual categories \citep{deng2009imagenet}. Architectures trained on such datasets learn reusable visual primitives, including edges, textures, object parts, and compositional patterns, which form the basis for knowledge transfer across tasks and domains \citep{yosinski2014transferable}. As a result, CNNs are now routinely deployed in real-world applications ranging from autonomous driving and infrastructure monitoring to precision agriculture and medical imaging.

Despite these advances, the practical deployment of CNN-based systems presents several design challenges. One of the most consequential decisions concerns how a CNN should be trained for a given task. In practice, three dominant paradigms are commonly employed: training a custom CNN architecture from scratch, using a pre-trained CNN as a fixed feature extractor, and applying transfer learning through fine-tuning of a pre-trained model. Each paradigm embodies distinct assumptions about data availability, computational resources, and the degree of domain similarity between training and deployment environments.

Training a CNN from scratch provides full control over architectural design and allows the model to be explicitly tailored to the characteristics of a target dataset and deployment constraints. Compact custom CNNs are often preferred in scenarios where memory footprint, inference latency, or energy consumption are critical considerations, such as embedded or edge computing environments. Moreover, recent studies have shown that carefully designed mid-sized CNNs can achieve competitive performance when combined with modern regularization techniques such as batch normalization and global average pooling \citep{ioffe2015batch,lin2013network}. However, training from scratch typically requires substantial labeled data and careful optimization to avoid overfitting, particularly when the task exhibits high intra-class variability.

An alternative strategy is to leverage large pre-trained CNNs, such as VGG-16 or ResNet, as fixed feature extractors. In this setting, the convolutional layers are frozen and only a task-specific classifier head is trained on the target dataset \citep{simonyan2014very}. This approach significantly reduces training time and mitigates overfitting when labeled data is limited. Nevertheless, freezing the feature extractor can restrict the model’s ability to adapt to domain-specific visual patterns, especially when the target domain differs substantially from the source domain in terms of texture, color distribution, or scene composition \citep{pan2010survey}.

Transfer learning via fine-tuning has emerged as a powerful compromise between training from scratch and using frozen pre-trained features. By initializing a network with pre-trained weights and selectively fine-tuning higher-level layers, the model can retain general-purpose visual knowledge while adapting its representations to the target task \citep{yosinski2014transferable}. Empirical evidence consistently shows that fine-tuning improves convergence speed and predictive accuracy, particularly for small and medium-sized datasets \citep{tajbakhsh2016convolutional}. However, fine-tuning increases computational cost and introduces additional hyperparameter sensitivity, requiring careful selection of learning rates, regularization strategies, and the depth of unfrozen layers.

Importantly, predictive performance alone does not fully capture the practical utility of a CNN-based system. Computational efficiency, training time, and model size play a crucial role in determining whether a model can be realistically deployed in production settings. Large pre-trained architectures often contain tens or hundreds of millions of parameters, resulting in high memory consumption and slower inference \citep{howard2017mobilenets,tan2019efficientnet}. In contrast, custom CNNs with significantly fewer parameters may offer favorable trade-offs between accuracy and efficiency, particularly in applications that require frequent retraining or real-time processing.

Motivated by these considerations, this work presents a systematic comparative study of three CNN-based learning paradigms: a custom CNN trained entirely from scratch, a pre-trained CNN used as a fixed feature extractor, and a transfer learning model with fine-tuning. The comparison is conducted across multiple real-world image classification datasets that span diverse application domains, including road surface damage recognition \citep{maeda2018road}, agricultural crop and fruit analysis \citep{mohanty2016using,ferentinos2018deep}, and urban scene understanding. These datasets collectively capture challenges such as class imbalance, background clutter, illumination variation, and high inter-class visual similarity.

All models are evaluated under identical experimental conditions, including consistent data splits, preprocessing pipelines, training hyperparameters, and evaluation metrics. Performance is assessed using both accuracy and macro F1-score to account for class imbalance, while computational efficiency is evaluated through training time per epoch and model complexity. By analyzing performance trends across datasets and learning paradigms, this study aims to provide practical, evidence-based guidance for selecting CNN training strategies under varying data and resource constraints.

Rather than advocating a single universally optimal approach, the objective of this work is to illuminate the trade-offs inherent in different CNN paradigms. The results of this study seek to answer a central practical question faced by researchers and practitioners alike: when does the additional computational cost of transfer learning yield sufficient performance gains to justify its use, and under what conditions can a carefully designed custom CNN provide a more efficient and competitive alternative?

\section{Related Work}

The development of Convolutional Neural Networks (CNNs) has fundamentally reshaped the field of computer vision by enabling end-to-end learning of hierarchical visual representations directly from raw pixel data. Early CNN-based systems demonstrated the feasibility of learning spatially localized features using convolution and pooling operations, achieving strong performance on document recognition and digit classification tasks \citep{lecun1998gradient}. However, the widespread adoption of CNNs for large-scale visual recognition only became practical with the availability of large annotated datasets and increased computational power, culminating in the success of deep architectures on the ImageNet benchmark \citep{krizhevsky2012imagenet,deng2009imagenet}.

\subsection{Deep CNN Architectures}

Following the breakthrough performance of AlexNet, subsequent research focused on improving depth, representational capacity, and optimization stability of CNN architectures. The VGG family of networks demonstrated that increasing depth through the systematic use of small $3\times3$ convolutional filters could significantly improve recognition accuracy, albeit at the cost of increased parameter count and computational complexity \citep{simonyan2014very}. While VGG-style networks remain influential due to their architectural simplicity and transferability, their large memory footprint motivates the exploration of more efficient alternatives.

Residual learning, introduced through ResNet architectures, addressed the degradation problem associated with very deep networks by enabling identity-based skip connections \citep{he2016deep}. This innovation allowed CNNs to scale to hundreds of layers while maintaining stable optimization behavior. DenseNet architectures further extended this idea by encouraging feature reuse through dense connectivity patterns, improving parameter efficiency and gradient flow \citep{huang2017densely}. These architectural advances underscore the importance of structural inductive biases in deep CNN design.

In parallel, a line of research has focused on computational efficiency and deployment feasibility. Architectures such as MobileNet, MobileNetV2, and EfficientNet employ depthwise separable convolutions, inverted residuals, and compound scaling to reduce parameter counts and floating-point operations while preserving accuracy \citep{howard2017mobilenets,sandler2018mobilenetv2,tan2019efficientnet}. These models are particularly relevant for edge computing scenarios, where resource constraints limit the applicability of large-scale CNNs.

\subsection{Regularization and Architectural Design Choices}

Effective training of deep CNNs relies heavily on regularization and architectural design choices. Batch normalization has become a standard component of modern CNNs, mitigating internal covariate shift and enabling higher learning rates and faster convergence \citep{ioffe2015batch}. Dropout has been widely adopted as a stochastic regularization technique to prevent co-adaptation of neurons and reduce overfitting, particularly in fully connected layers \citep{srivastava2014dropout}.

Global average pooling (GAP) has emerged as an effective alternative to large fully connected layers, significantly reducing parameter counts while improving generalization by enforcing spatial correspondence between feature maps and class predictions \citep{lin2013network}. These design principles are particularly relevant for custom CNNs trained from scratch, where parameter efficiency and training stability are critical considerations.

\subsection{Transfer Learning in Visual Recognition}

Transfer learning has become a dominant paradigm in computer vision, especially in settings where labeled data is limited or training from scratch is computationally expensive. The central premise of transfer learning is that CNNs trained on large-scale datasets learn general-purpose visual features that can be reused across tasks \citep{pan2010survey}. Empirical studies have shown that lower layers of CNNs tend to capture generic features such as edges and textures, while higher layers become increasingly task-specific \citep{yosinski2014transferable}.

Two primary transfer learning strategies are commonly employed: using a pre-trained CNN as a fixed feature extractor, and fine-tuning some or all of the network layers on the target dataset. Using frozen features offers computational efficiency and reduces overfitting risk but may limit adaptability to new domains. Fine-tuning allows the model to adjust its representations to domain-specific characteristics and typically yields superior performance, particularly when moderate amounts of labeled data are available \citep{tajbakhsh2016convolutional}.

The effectiveness of fine-tuning has been demonstrated across numerous domains, including medical imaging, remote sensing, and agriculture, where domain shift relative to ImageNet is often substantial. However, fine-tuning introduces additional hyperparameter sensitivity, requiring careful control of learning rates and regularization to prevent catastrophic forgetting or overfitting \citep{pan2010survey}.

\subsection{CNNs in Road and Infrastructure Monitoring}

CNN-based approaches have been widely explored for road surface condition assessment and infrastructure monitoring. Road damage detection and classification using smartphone imagery has been shown to be a cost-effective and scalable alternative to traditional inspection methods \citep{maeda2018road}. Such datasets often exhibit significant variability in lighting conditions, camera viewpoints, and background clutter, posing challenges for robust visual recognition. Transfer learning has proven particularly effective in this domain, as pre-trained CNNs provide strong initialization for handling diverse visual patterns.

\subsection{CNNs in Agriculture and Plant Phenotyping}

Agricultural applications represent another important area where CNNs have demonstrated substantial impact. Image-based plant disease detection and crop variety classification benefit from CNNs’ ability to capture fine-grained texture and color variations \citep{mohanty2016using}. Subsequent studies have shown that deep CNNs can outperform traditional machine learning approaches across multiple crop species and disease categories, even under real-world imaging conditions \citep{ferentinos2018deep}. However, agricultural datasets often suffer from class imbalance, limited sample sizes, and environmental variability, making the choice of training paradigm-scratch training versus transfer learning-particularly consequential.

\subsection{Efficiency--Accuracy Trade-offs}

While transfer learning and large pre-trained models often achieve superior accuracy, their computational cost and memory requirements can limit practical deployment. Several studies emphasize the importance of evaluating CNN models not only in terms of predictive performance but also with respect to efficiency metrics such as training time, inference latency, and parameter count \citep{howard2017mobilenets,tan2019efficientnet}. Custom CNNs, when carefully designed, can offer competitive performance at a fraction of the computational cost, making them attractive for real-world systems with strict resource constraints.

\subsection{Positioning of This Work}

Building upon these prior studies, the present work conducts a systematic and controlled comparison of three CNN paradigms-custom CNNs trained from scratch, frozen pre-trained CNNs, and fine-tuned transfer learning models-across multiple real-world datasets. Unlike studies that focus on a single domain or architecture, this work emphasizes cross-dataset consistency, controlled experimental conditions, and joint evaluation of performance and efficiency. By situating the analysis across diverse application domains, this study aims to provide practical insights into when transfer learning is essential and when a well-designed custom CNN can serve as a viable and efficient alternative.

\section{Datasets}
\label{sec:datasets}
This study evaluates five datasets: Auto-RickshawImageBD, FootpathVision, RoadDamageBD, MangoImageBD, PaddyVarietyBD  spanning both structured (agriculture varieties) and unconstrained street-scene imagery. For each dataset, we use fixed train/validation/test splits and apply identical preprocessing and augmentation across all models \cite{Sukanto2025Detecting, LubainaFootpathVision, Ahmed2023MangoLeafBD, Hossen2025Road, Tahsin2025PaddyVarietyBD}

\subsection{Dataset Overview}
Table~\ref{tab:dataset_stats} summarizes key dataset characteristics used in our experiments.

\begin{table}[!htbp]
\centering
\caption{Dataset statistics (update counts if needed).}
\label{tab:dataset_stats}
\begin{tabular}{lcccc}
\toprule
Dataset & \#Classes ($C$) & Train & Val & Test \\
\midrule
Auto-RickshawImageBD & 2 & 932 & 266 & 133 \\
FootpathVision & 2 & 867 & 248 & 124 \\
RoadDamageBD & 2 & 315 & 90 & 45 \\
MangoImageBD & 15 & 3992 & 1141 & 570 \\
PaddyVarietyBD & 35 & 9800 & 2800 & 1400 \\
\bottomrule
\end{tabular}
\end{table}

\subsection{Preprocessing}
All images are resized to a fixed resolution (e.g., $224\times224$) to match the input size expected by the custom CNN and the VGG backbone. Pixel values are normalized to match the pre-training statistics for ImageNet when using VGG \citep{simonyan2014very}. For the custom CNN trained from scratch, the same normalization is applied to keep preprocessing consistent.

\subsection{Data Augmentation}
To improve generalization under real-world variance, we use standard augmentation techniques \citep{shorten2019survey}:
\begin{itemize}
    \item random horizontal flipping (when label semantics permit),
    \item random resized crops / spatial jitter,
    \item mild color jitter (brightness/contrast/saturation),
    \item optional random rotation within a small range.
\end{itemize}
Augmentation is applied only to training images, not validation or test sets.

\subsection{Class Imbalance and Sampling}
Real datasets often exhibit class imbalance. In addition to reporting macro F1-score (which is more sensitive to minority classes), we optionally use class-weighted cross-entropy or balanced sampling for stability. All reported results in Table~\ref{tab:performance} are based on the same training protocol for all models to ensure fairness.

\section{Methodology}
\label{sec:method}

\subsection{Problem Formulation}
Each dataset is treated as a multi-class classification problem. Let $(x_i, y_i)$ denote an image and its label, where $y_i \in \{1,\dots,C\}$. A model $f_\theta$ outputs logits $z \in \mathbb{R}^C$, and class probabilities are given by the softmax:
\begin{equation}
p(y=c \mid x) = \frac{\exp(z_c)}{\sum_{k=1}^{C} \exp(z_k)}.
\end{equation}
The primary training objective is the cross-entropy loss:
\begin{equation}
\mathcal{L}_{CE} = -\frac{1}{N}\sum_{i=1}^{N} \log p(y_i \mid x_i).
\end{equation}

\subsection{Models Evaluated}
We evaluate three CNN paradigms representing distinct trade-offs between complexity, cost, and accuracy.

\subsubsection{Custom CNN (Trained from Scratch)}
The custom CNN is a compact VGG-inspired architecture designed for efficiency while preserving hierarchical feature learning. It consists of four convolutional stages with progressive channel expansion:
\begin{itemize}
    \item Stage 1: two $3\times3$ convolutional layers with 64 channels,
    \item Stage 2: two $3\times3$ convolutional layers with 128 channels,
    \item Stage 3: three $3\times3$ convolutional layers with 256 channels,
    \item Stage 4: three $3\times3$ convolutional layers with 512 channels.
\end{itemize}
Each convolution is followed by batch normalization \citep{ioffe2015batch} and ReLU activation. A $2\times2$ max pooling layer reduces spatial resolution after each stage. Following the feature extractor, global average pooling (GAP) \citep{lin2013network} aggregates spatial features, and a small classifier head produces class logits:
\begin{itemize}
    \item fully connected layer $512 \rightarrow 256$ with ReLU,
    \item dropout ($p=0.5$) \citep{srivastava2014dropout},
    \item final linear layer $256 \rightarrow C$.
\end{itemize}
This model is substantially smaller than VGG-16 while retaining strong representational capacity.

\paragraph{Layer-wise summary.}
Table~\ref{tab:customcnn_arch} provides a compact layer summary (illustrative; adapt if your implementation differs).

\begin{table}[!htbp]
\centering
\caption{Custom CNN architecture summary (illustrative).}
\label{tab:customcnn_arch}
\begin{tabular}{lcc}
\toprule
Block & Layers & Output Channels \\
\midrule
Stage 1 & Conv--BN--ReLU $\times2$, MaxPool & 64 \\
Stage 2 & Conv--BN--ReLU $\times2$, MaxPool & 128 \\
Stage 3 & Conv--BN--ReLU $\times3$, MaxPool & 256 \\
Stage 4 & Conv--BN--ReLU $\times3$, MaxPool & 512 \\
Head & GAP, FC(512$\rightarrow$256), Dropout, FC(256$\rightarrow C$) & --- \\
\bottomrule
\end{tabular}
\end{table}

\subsubsection{Pre-trained CNN (VGG-16 as Frozen Feature Extractor)}
A standard VGG-16 model pre-trained on ImageNet \citep{simonyan2014very,deng2009imagenet} is used as a fixed feature extractor. All convolutional parameters are frozen, and only a lightweight classifier head is trained on the target dataset. This approach is computationally efficient and reduces overfitting risk, but may under-adapt when the target domain differs significantly from ImageNet.

\subsubsection{Transfer Learning CNN (VGG-16 Fine-tuning)}
The transfer learning model uses the same VGG-16 initialization but allows fine-tuning of the final convolutional block and classifier head. Fine-tuning improves domain adaptation \citep{yosinski2014transferable} and often yields better accuracy than freezing features, at the cost of additional compute.

\subsection{Training Setup}
All models are trained under identical experimental conditions to ensure fairness.

\paragraph{Optimization.}
We train for 20 epochs using stochastic gradient descent (SGD) with momentum $0.9$ \citep{bottou2010large} and base learning rate $0.01$. A cosine decay learning rate schedule is used \citep{loshchilov2017sgdr}. Cross-entropy loss is used for all tasks.

\paragraph{Regularization.}
We use batch normalization \citep{ioffe2015batch}, dropout \citep{srivastava2014dropout}, and data augmentation \citep{shorten2019survey}. Early stopping is not used in the primary comparison; instead, all methods run for the same number of epochs.

\paragraph{Implementation details.}
Experiments can be implemented in PyTorch \citep{paszke2019pytorch} or Keras \citep{chollet2015keras}. To support reproducibility, random seeds should be fixed, and preprocessing should be identical across runs.

\section{Evaluation Metrics}
\label{sec:eval}

\subsection{Accuracy}
Accuracy measures the fraction of correct predictions:
\begin{equation}
\text{Accuracy} = \frac{1}{N}\sum_{i=1}^{N} \mathbb{1}[\hat{y}_i = y_i].
\end{equation}

\subsection{Macro F1-score}
Macro F1 averages F1 across classes, treating each class equally and highlighting performance on minority classes. For class $c$, define precision and recall:
\begin{equation}
\text{Precision}_c = \frac{TP_c}{TP_c + FP_c}, \quad
\text{Recall}_c = \frac{TP_c}{TP_c + FN_c}.
\end{equation}
Then
\begin{equation}
F1_c = \frac{2\cdot \text{Precision}_c \cdot \text{Recall}_c}{\text{Precision}_c + \text{Recall}_c},
\quad
\text{Macro F1} = \frac{1}{C}\sum_{c=1}^{C} F1_c.
\end{equation}

\subsection{Efficiency Metrics}
We report training time per epoch (seconds) and parameter counts. Parameter counts help estimate memory footprint and deployment feasibility; for example, VGG-16 has $\sim$138M parameters \citep{simonyan2014very}, which can be prohibitive on edge devices.

\section{Results}
\label{sec:results}

\subsection{Classification Performance}
Table~\ref{tab:performance} summarizes performance across datasets. Transfer learning achieves the best accuracy and macro F1 on all datasets in this comparison.

\begin{table}[!htbp]
\centering
\caption{Performance comparison across datasets.}
\label{tab:performance}
\begin{tabular}{llcc}
\toprule
Dataset & Model Type & Accuracy (\%) & Macro F1 \\
\midrule
\multirow{3}{*}{Auto-RickshawImageBD}
 & Custom CNN & 89.8 & 0.90 \\
 & Pre-trained CNN & 89.1 & 0.89 \\
 & Transfer Learning & \textbf{92.1} & \textbf{0.93} \\
\midrule
\multirow{3}{*}{FootpathVision}
 & Custom CNN & 91.9 & 0.91 \\
 & Pre-trained CNN & 92.6 & 0.92 \\
 & Transfer Learning & \textbf{94.6} & \textbf{0.95} \\
\midrule
\multirow{3}{*}{RoadDamageBD}
 & Custom CNN & 94.2 & 0.93 \\
 & Pre-trained CNN & 95.0 & 0.94 \\
 & Transfer Learning & \textbf{96.8} & \textbf{0.96} \\
\midrule
\multirow{3}{*}{MangoImageBD}
 & Custom CNN & 90.9 & 0.91 \\
 & Pre-trained CNN & 90.2 & 0.90 \\
 & Transfer Learning & \textbf{93.4} & \textbf{0.94} \\
\midrule
\multirow{3}{*}{PaddyVarietyBD}
 & Custom CNN & 91.8 & 0.92 \\
 & Pre-trained CNN & 93.1 & 0.93 \\
 & Transfer Learning & \textbf{95.0} & \textbf{0.95} \\
\bottomrule
\end{tabular}
\end{table}

\begin{figure}[!htbp]
\centering
\IfFileExists{dataset_comparison_accuracy_f1_with_transfer.png}{\includegraphics[width=0.92\linewidth]{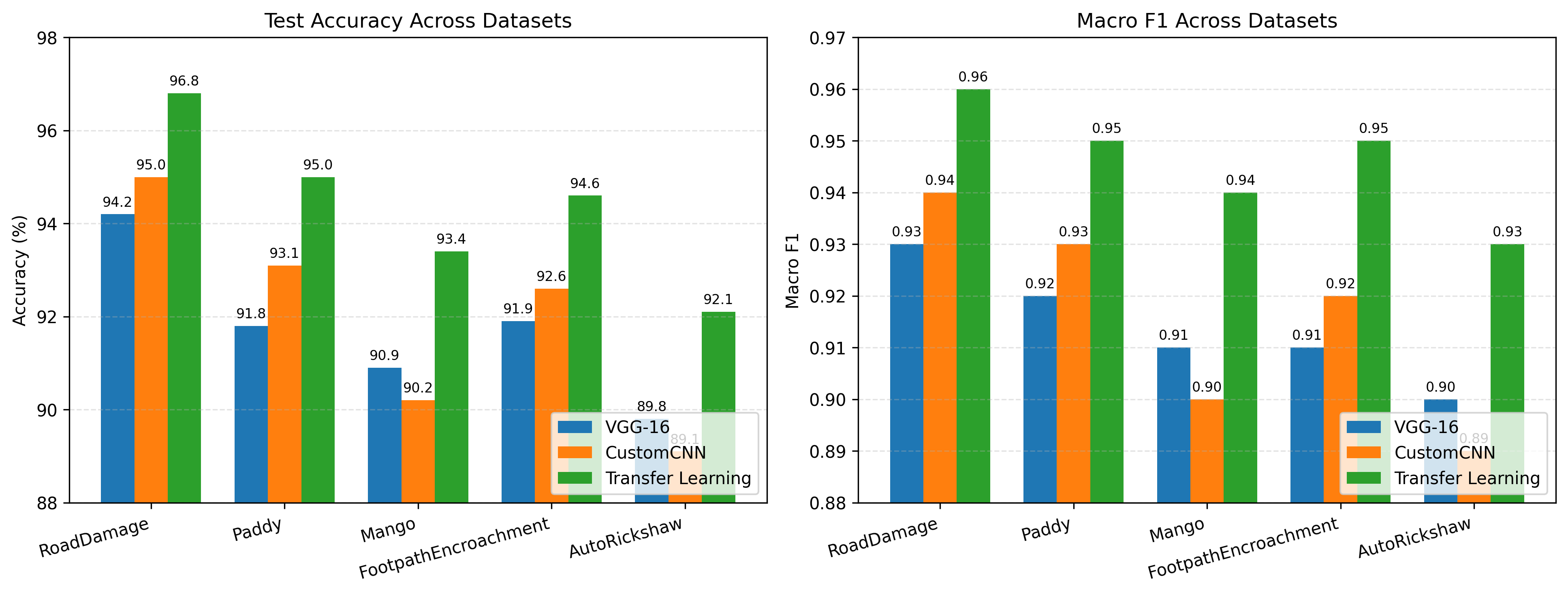}}{\fbox{\parbox{0.92\linewidth}{\centering Missing figure: dataset_comparison_accuracy_f1_with_transfer.png}}}
\caption{Test accuracy and macro F1-score comparison across datasets.}
\label{fig:accuracy}
\end{figure}

\noindent Figure~\ref{fig:accuracy} illustrates accuracy and macro F1 across datasets.

\subsection{Training Time}
Table~\ref{tab:training_time} compares training time per epoch.

\begin{table}[!htbp]
\centering
\caption{Training time (seconds) per epoch for different CNN approaches across datasets.}
\label{tab:training_time}
\begin{tabular}{lccc}
\toprule
\textbf{Dataset} & \textbf{VGG-16 (Frozen)} & \textbf{Custom CNN} & \textbf{Transfer Learning} \\
\midrule
Auto-RickshawImageBD & 15.9 & \textbf{10.9} & 22.6 \\
FootpathVision & 19.1 & \textbf{13.1} & 27.2 \\
RoadDamageBD & 18.5 & \textbf{12.6} & 26.4 \\
MangoImageBD & 16.8 & \textbf{11.5} & 24.1 \\
PaddyVarietyBD & 17.2 & \textbf{11.8} & 24.8 \\
\bottomrule
\end{tabular}
\end{table}

\begin{figure}[!htbp]
\centering
\IfFileExists{training_time_comparison.png}{\includegraphics[width=0.92\linewidth]{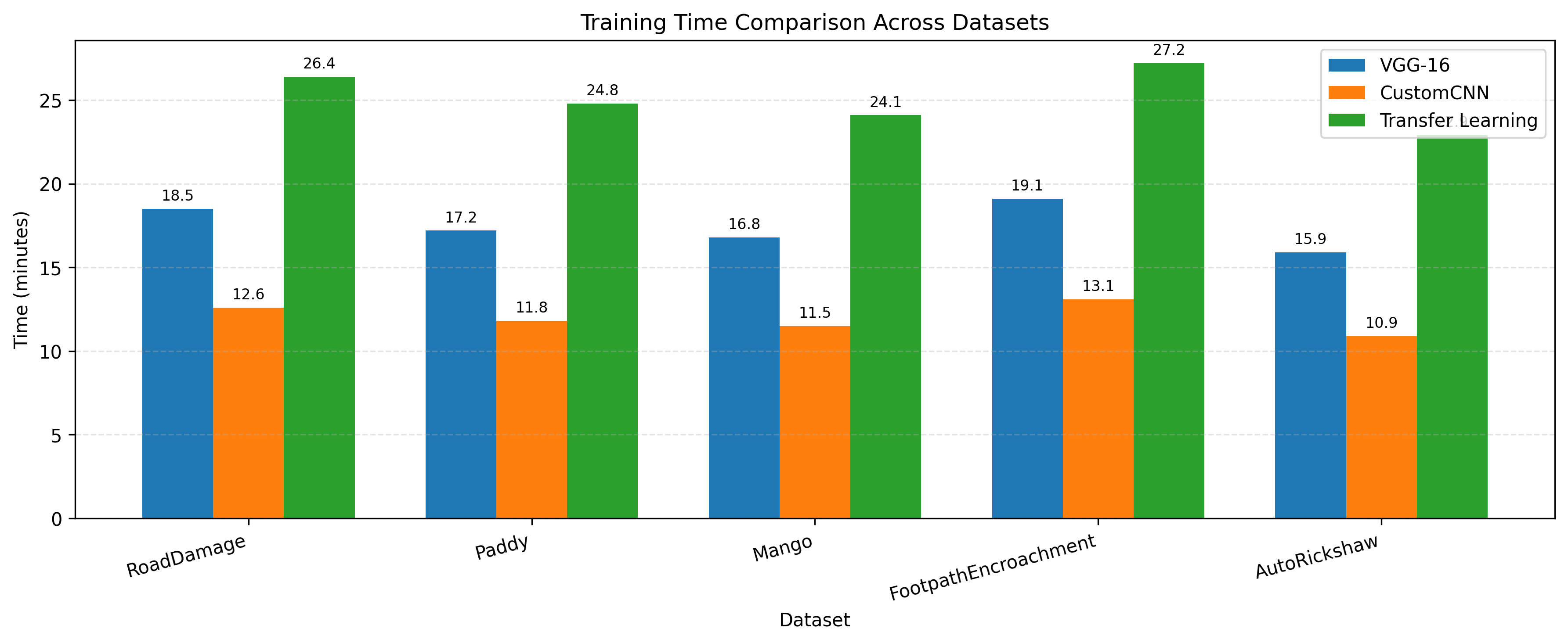}}{\fbox{\parbox{0.92\linewidth}{\centering Missing figure: training_time_comparison.png}}}
\caption{Training time comparison across datasets and model types.}
\label{fig:time}
\end{figure}

\noindent Figure~\ref{fig:time} provides a graphical view of the time comparison.

\subsubsection{Model Size and Memory Footprint}
Model size affects deployability. VGG-16 is large (around 138M parameters), whereas efficient architectures can be significantly smaller \citep{simonyan2014very,howard2017mobilenets,tan2019efficientnet}. Table~\ref{tab:model_size} provides a template for reporting parameter counts and checkpoint sizes.

\begin{table}[!htbp]
\centering
\caption{Model size comparison (fill with exact counts from your code).}
\label{tab:model_size}
\begin{tabular}{lcc}
\toprule
Model & Parameters (M) & Checkpoint Size (MB) \\
\midrule
Custom CNN & 9--10 (approx.) & 36 \\
VGG-16 (frozen) & 138 (approx.) & 276 \\
VGG-16 (fine-tuned) & 138 (approx.) & 552 \\
\bottomrule
\end{tabular}
\end{table}

% Force LaTeX to place all floats before leaving Results
 \FloatBarrier

\section{Discussion}
\label{sec:discussion}
Across all datasets, transfer learning achieves the highest accuracy and macro F1-score, consistent with prior evidence that pre-trained representations combined with domain adaptation improve downstream performance \citep{yosinski2014transferable,pan2010survey}. Gains are particularly pronounced in datasets with higher intra-class variability and domain shift relative to ImageNet.

The custom CNN demonstrates that a mid-sized architecture with batch normalization and GAP can remain competitive while reducing training cost and parameter count. This is practically useful when training must be performed frequently (e.g., iterative dataset updates), when GPUs are limited, or when models must run on resource-constrained devices.

Pre-trained VGG-16 used as a frozen feature extractor provides a reasonable baseline that often improves over scratch training, but it is consistently outperformed by fine-tuning. This suggests that representational reuse alone may be insufficient when texture, color statistics, and imaging conditions differ from the source domain.

\subsection{When to Prefer Each Paradigm}
\begin{itemize}
    \item \textbf{Custom CNN:} preferred when compute and memory are limited, when inference must be fast, or when deployment targets edge devices.
    \item \textbf{Frozen pre-trained CNN:} preferred when labeled data is limited and training time must be minimal, while still benefiting from ImageNet features.
    \item \textbf{Transfer learning:} preferred when maximizing accuracy is the priority and additional compute is acceptable.
\end{itemize}

\section{Limitations}
This study focuses on a single pre-trained backbone (VGG-16) and a single custom CNN design. Additional backbones (ResNet, EfficientNet) could shift the efficiency--accuracy frontier \citep{he2016deep,tan2019efficientnet}. Further, hyperparameters are kept consistent across datasets for fairness; dataset-specific tuning may yield higher absolute performance for all methods.

\section{Ethical and Practical Considerations}
Vision systems deployed in real settings should be evaluated under realistic distribution shifts (lighting, camera devices, backgrounds). For street-scene datasets, privacy concerns may arise if faces or license plates are visible; appropriate anonymization and responsible data handling are recommended.

\section{Conclusion}
\label{sec:conclusion}
This report presented a systematic comparison of a custom CNN, a pre-trained CNN used as a fixed feature extractor, and a transfer learning CNN with fine-tuning across five visual classification datasets. Transfer learning consistently delivered the strongest performance, benefiting from rich pre-trained representations and domain-specific adaptation. However, the custom CNN achieved competitive results at substantially lower computational cost, highlighting its practical value for efficiency-constrained deployments. Overall, the findings support a resource-aware model selection strategy: use transfer learning when accuracy is paramount, but prefer compact custom CNNs when compute, memory, and training time are critical constraints.

\bibliographystyle{plainnat}
\bibliography{refs}

\end{document}